\pdfoutput=1
\documentclass[11pt]{article}

\usepackage{acl}

\usepackage{times}
\usepackage{latexsym}
\usepackage{graphicx}

\usepackage{array}
\usepackage{mathtools}
\usepackage{subcaption}
\usepackage{pbox}
\usepackage{makecell}
\usepackage{float}
\usepackage{tabularx}
\usepackage{amsfonts}
\usepackage[group-separator={,},group-minimum-digits={3}]{siunitx}
\newcolumntype{P}[1]{>{\centering\arraybackslash}p{#1}}
\usepackage{scalerel,xparse}

\usepackage[T1]{fontenc}
\usepackage[utf8]{inputenc}

\usepackage{microtype}

\title{Opponent Modeling in Negotiation Dialogues\\by Related Data Adaptation}

\author{Kushal Chawla$^{1}$\hspace{0.3cm}Gale M. Lucas$^1$\hspace{0.3cm} \textbf{Jonathan May}$^2$\hspace{0.3cm} \textbf{Jonathan Gratch}$^1$ \\
University of Southern California, Los Angeles, USA \\
$^1$\texttt{\{chawla,lucas,gratch\}@ict.usc.edu}
\\$^2$\texttt{jonmay@isi.edu}}

\begin{document}
\maketitle
\begin{abstract}
Opponent modeling is the task of inferring another party's mental state within the context of social interactions. In a multi-issue negotiation, it involves inferring the relative importance that the opponent assigns to each issue under discussion, which is crucial for finding high-value deals. A practical model for this task needs to infer these priorities of the opponent on the fly based on partial dialogues as input, without needing additional annotations for training. In this work, we propose a ranker for identifying these priorities from negotiation dialogues. The model takes in a partial dialogue as input and predicts the priority order of the opponent. We further devise ways to adapt related data sources for this task to provide more explicit supervision for incorporating the opponent's preferences and offers, as a proxy to relying on granular utterance-level annotations. We show the utility of our proposed approach through extensive experiments based on two dialogue datasets. We find that the proposed data adaptations lead to strong performance in zero-shot and few-shot scenarios. Moreover, they allow the model to perform better than baselines while accessing fewer utterances from the opponent. We release our code to support future work in this direction: \url{https://github.com/kushalchawla/opponent-modeling}.
\end{abstract}

\section{Introduction}
Negotiations are key to our everyday interactions such as allocating available resources, salary decisions, business deals, and legal proceedings. The ability to effectively negotiate is also critical for automated systems deployed in complex social scenarios~\cite{gratch2015negotiation}. This enables these automated systems to engage in strategic conversations~\cite{leviathan2018google} and also assists in pedagogy by making social skills training more accessible~\cite{johnson2019intelligent}.

\begin{figure}[t!]
\centering
\includegraphics[width=\linewidth]{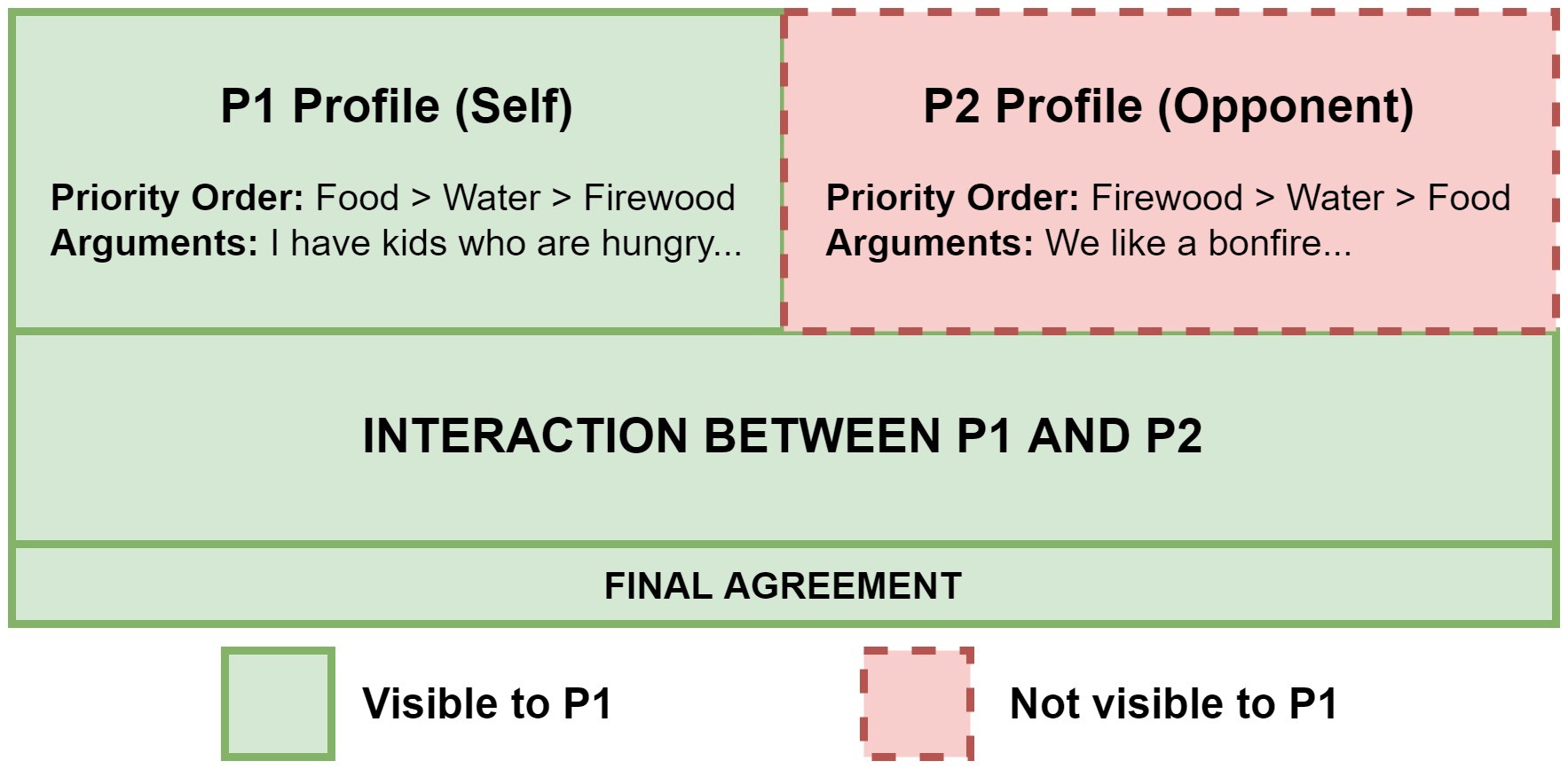}
\caption{A simplified view of a multi-issue negotiation based on the scenario in CaSiNo~\cite{chawla2021casino}. The negotiation involves $3$ issues: Food, Water, and Firewood, each with $3$ items that must be divided among the two players. From the perspective of player P1, the task of opponent modeling considered in this work involves inferring the priority order of the opponent P2 from the interaction between the two.}
\label{fig:intro}
\end{figure}

Consider the scenario presented in Figure \ref{fig:intro}. Two participants role-play as campsite neighbors and engage in a multi-issue negotiation~\cite{fershtman1990importance} over three issues: food, water, and firewood~\cite{chawla2021casino}. Each negotiator has their own priority order depending on the relative importance assigned to each issue. The goal of the negotiation is to divide the available quantities of food, water, and firewood packages, such that each package is assigned to exactly one of the players in the final agreement.

The priority order of the opponent is typically unknown to negotiators beforehand, and can only be inferred based on the interaction between the two. Prior work argues that understanding what one's opponent wants is one of the key aspects of successful negotiations~\cite{baarslag2013predicting}. An accurate \textit{model} of the opponent can enable a dialogue system to roll out offers that work for both parties, which has implications on both its objective performance such as the final points scored from the agreed deal, and the subjective performance such as opponent's satisfaction and affinity for the dialogue system. This can also aid in pedagogy by allowing the system to provide concrete feedback to students who fail to incorporate the priorities of their opponents~\cite{johnson2019assessing}. Discovering these priorities from an interaction with an opponent is usually referred to as \textit{Opponent modeling} in the context of multi-issue negotiations.

Information about an opponent's priorities can primarily be gathered from their preference and offer statements~\cite{nazari2015opponent}. Sharing preferences by explicitly mentioning \textit{`We need water'} or more implicitly - \textit{`We like to go on runs'} can provide information that water is of high priority to the negotiator. Further, offers such as \textit{`I would like two food items and one water'} can imply that food is of a higher priority than water.

Building techniques for opponent modeling that are useful in realistic chat-based negotiations poses several key challenges: \textbf{1)} It is non-trivial to directly use counting-based methods on these preference and offer statements, which are common in prior work that does not use natural language, such as agent-agent negotiations~\cite{williams2012iamhaggler} and human-agent negotiations based on button clicks~\cite{mell2017grumpy}, \textbf{2)} To alleviate this problem for language-based interactions, prior work has resorted to gathering additional utterance-level annotations to convert the desirable information into a more structured format, that can then be used with counting methods~\cite{nazari2015opponent}. However, this approach remains expensive, requires expertise, and hurts generalizability. Further, these annotations are unavailable for systems that are deployed to end users, needing a separate NLU module which can potentially lead to error propagation in the downstream dialogue system pipeline, and \textbf{3)} Some real-world applications require the system to guess the opponent's priorities with only partial dialogue so as to inform the future decision process of the system - a scenario which has not been well explored in prior works.

To address these challenges, we propose a transformer-based~\cite{vaswani2017attention} hierarchical ranker for opponent modeling in negotiation dialogues. Our model takes a partial dialogue as input and guesses the opponent's priority order. Instead of relying on utterance-level discourse information, we devise simple and effective ways to project related data sources to this task. As opposed to multi-task learning which typically involves task-agnostic and task-specific parameters and back-to-back fine-tuning procedures that suffer from catastrophic forgetting issues, our adaptations augment the training data available to the model, allowing end-to-end joint learning and parameter sharing. We summarize our contributions below:
\begin{enumerate}
    \item We formulate opponent modeling as a ranking task (Section \ref{sec:problem-formulation}) and propose a transformer-based model that can be trained directly on partial dialogues using a pairwise margin ranking loss (Section \ref{sec:methodology}).
    \item To better capture the opponent preferences and offers, we devise methods to adapt related data sources, resulting in more labeled data for training (Section \ref{sec:methodology}).
    \item For a comprehensive evaluation that serves multiple downstream applications, we propose three evaluation metrics for this task (Section \ref{sec:expt-design}). Our experiments are based on two dialogue datasets in English: CaSiNo~\cite{chawla2021casino} and DealOrNoDeal~\cite{lewis2017deal}, showing the utility of the proposed methodology with complete or partial dialogue as input in full, few-shot, and zero-shot scenarios (Section \ref{sec:results}).
    \item We compare our best-performing model to a human expert, discussing common errors to guide future work (Section \ref{sec:results}), and laying out the implications for research in human-machine negotiations (Section \ref{sec:broader-ethics}).
\end{enumerate}

\begin{figure*}[th]
\centering
 \includegraphics[width=\linewidth]{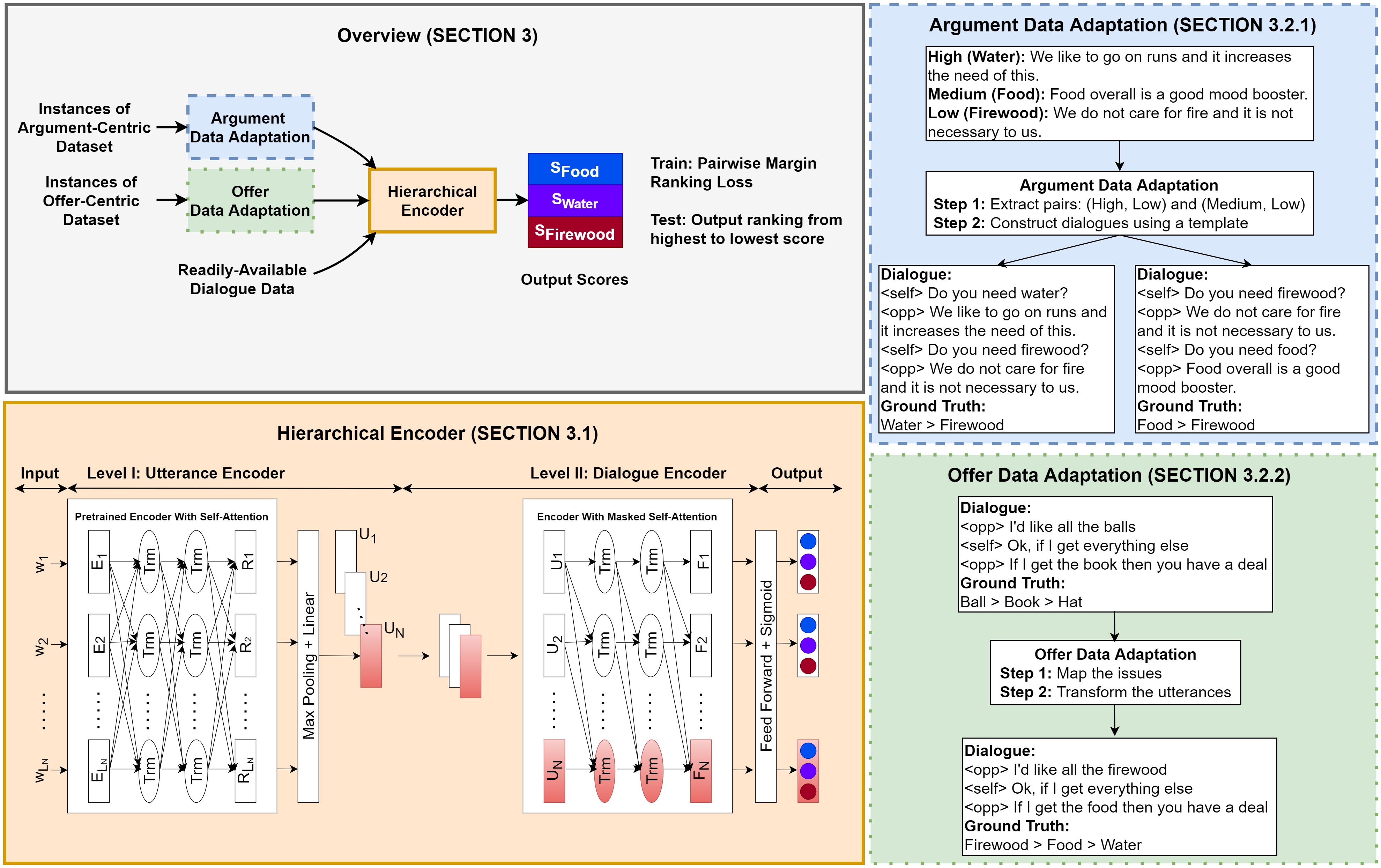}
\caption{Our proposed methodology for opponent modeling in negotiation dialogues. The approach involves three main components: Section 3.1 describes our core hierarchical encoder that takes in a partial dialogue and outputs the opponent priority order after seeing each utterance, Section 3.2.1 covers the adaptation of an argument-centric dataset (CA data) targeted towards better modeling the preference statements of the opponent, and Section 3.2.2 describes the adaptation of an offer-centric dataset (DND data) targeted towards the offer statements of the opponent.}
\label{fig:architecture}
\end{figure*}

\section{Problem Formulation}
\label{sec:problem-formulation}
Consider a negotiation $C$ between two parties over $m$ issues. We define the problem from the \textit{perspective} of a specific negotiator (referred to as \textit{self}, hereafter), and aim to predict the priority order of the \textit{opponent} (see Figure \ref{fig:intro}). Assume that $C$ contains an alternating sequence of $N$ utterances between the negotiator self $S$ and the opponent $O$. The partial interaction is $C_k$, which is obtained after $S$ observes $k$ utterances from the opponent.\footnote{$C_k$ will contain either $2k$ or $2k-1$ utterances, depending on who starts the conversation.} The goal is to build the model $M$, with $Y_O = M(C_k)$, where $Y_O$ is the desired priority order of the opponent. In our experiments, we consider metrics that measure the performance for the complete dialogue and for different values of $k$ (Section \ref{sec:results}).

\section{Methodology}
\label{sec:methodology}
We present our approach in Figure \ref{fig:architecture}, which contains three main components: a hierarchical core model that takes in a partial dialogue and outputs the desired ranking order, and two modules for data adaptation that are designed to better model the preference and offer statements of the opponent. We first describe our core model, assuming a general input, and then describe the proposed data augmentation techniques.

\subsection{Hierarchical Encoder}
Our encoder (orange segment from Figure \ref{fig:architecture}) uses two levels to build contextual utterance representations, which are then used to output a score for each of the $m$ issues, representing the ranking order among them.

\noindent\textbf{Utterance Encoder:} First, a sentence-level module (Level I) encodes each utterance U$_j$ = $[w_1, w_2, \dots, w_{L_j}]$ separately. We prepend the utterances with a special token to indicate the author: <self> or <opp>. To encode a contextually-rich representation, our level I encoder uses pretrained language models~\cite{devlin2019bert, liu2019roberta}, given their success across a wide range of NLP tasks, especially in low resource settings on similar NLU tasks ~\cite{balaraman2021recent}. For each utterance $U_j$, the pretrained model first embeds the input words into the embedding matrix $E \in \mathbb{R}^{L_j \times d}$. After passing through the encoding layers, the pretrained model outputs $d$-dimensional word representations $R \in \mathbb{R}^{L_j \times d}$. Finally, this is followed by pooling to obtain the utterance representation $U_j \in \mathbb{R}^{d}$. The Level I output is essentially the conversation matrix $U \in \mathbb{R}^{N \times d}$, which is obtained after processing all the input utterances.

\noindent\textbf{Dialogue Encoder:} At Level II, we use a transformer block with masked self-attention~\cite{vaswani2017attention}. Self-attention enables efficient interactions for encoding partial conversations. A target utterance is only allowed to use the information from previously-seen utterances, which is accomplished by masking all the future utterances in the dialogue. In a single transformer layer, each target utterance \textit{query} simultaneously assesses and encodes the information from all the unmasked \textit{key} utterances, resulting in a contextualized representation of each utterance - the matrix $F \in \mathbb{R}^{N \times d}$.

\noindent\textbf{Output Layers:} Finally, a feed-forward network acts on $F$ to output an $m$-dimensional representation for each utterance. This represents the scores for each of the issues that the model is trying to rank. We then apply the sigmoid operation to constrain each score between $0$ and $1$, resulting in the output $O \in \mathbb{R}^{N \times m}$.

In comparison to text ranking tasks where the set of items that are being ranked is large and can be dynamic, the set of issues in realistic multi-issue negotiations is usually small and fixed. Hence, we predict the scores for each of these issues together, unlike text ranking literature where each item is ranked separately~\cite{yates2021pretrained}.

\noindent\textbf{Training:} We employ the pairwise margin ranking loss to train our model in an end-to-end manner. The loss $\mathcal{L}_k$ after observing $k$ utterances from the opponent is defined as:
\begin{equation}
\label{eq:margin-loss}
    \mathcal{L}_k = \sum_{q=(q_1, q_2) \in Q} L_k(o_{q_1}^k, o_{q_2}^k, y_q),
\end{equation}
where $L_k$ is given by:
\begin{equation}
     L_k(o_{q_1}^k, o_{q_2}^k, y_q) = max(0, -y_q(o_{q_1}^k-o_{q_2}^k) + c).
\end{equation}
$Q$ represents the set of all possible pairs of issues. $o_{q_1}^k$ and $o_{q_2}^k$ are the scoress from the final layer of the hierarchical ranker after applying the sigmoid operation. $y_q$ captures the ground truth ranking between $q_1$ and $q_2$. $y_q$ is equal to $+1$ when $q_1$ should be ranked higher (has a larger score) than $q_2$ and it is kept as $-1$ otherwise. $c$ is the margin.

The objective of the ranking loss is to train the model to predict a \textit{higher} score for the issue that is ranked \textit{higher} by the ground truth priority order. A positive margin of $c$ ensures a nonzero loss if the score for the higher ranked item is \textit{not greater than or equal to its counterpart by $c$}, forcing the model to predict well-separated boundaries. We experimented with different values for $c$, concluding that a nonzero margin is necessary for any meaningful training. For the results presented in this paper, we set $c$ as $0.3$.

\noindent\textbf{Inference:} Once the model is trained, the predicted scores can be used to output the desired ranking order for a given input dialogue. The model simply outputs the ranking of the issues by ordering them in decreasing order of these predicted scores.

\noindent\textbf{Note on the loss formulation}: The pairwise ranking loss was chosen for its suitability and simplicity. However, other potential alternatives do exist. Since the number of issues is limited, one can remodel the prediction task as classification over all the possible orderings. However, this trivially does not capture that although two orderings can be wrong, one can be \textit{somewhat less} wrong than the other. Hence, a ranking loss is more suitable for giving a smoother signal to the model during training, leading to a better performance in our initial experiments. We also explored more complicated ranking loss functions and a sequence-to-sequence model to directly generate the sequence of issues in their correct ranking order~\cite{yates2021pretrained}. We instead found the pairwise ranking loss to be effective and simple for our approach in this paper that involves a limited set of issues and exploits partially-masked loss functions (Section \ref{sec:capture-preferences}). Regardless, we encourage future work to explore these other formulations as well depending on the task at hand.

\subsection{Data Adaptations}
The transformer model discussed above learns to rank the issues directly from the partial dialogue as input without any additional supervision. Although this approach performs reasonably well in our experiments, it ignores the observations made in prior work which have primarily relied on annotations for preference and offer statements for opponent modeling~\cite{nazari2015opponent}. This suggests that more explicit feedback for extracting information from preferences and offers is one avenue for improving the performance, especially in settings when the available dialogue data is scarce. Instead of gathering additional annotations, we devise alternate ways to better capture the preferences and offers in our hierarchical ranking model. We achieve this by adapting two additional data sources for this task, allowing the data to be directly added to the primary training dataset and enabling end-to-end parameter sharing between these related tasks.

\noindent\textbf{Datasets}: We leverage two datasets in this work: CaSiNo~\cite{chawla2021casino} and DealOrNoDeal~\cite{lewis2017deal}. As discussed before, CaSiNo is grounded in a camping scenario, containing negotiations over three issues: \textit{food}, \textit{water}, and \textit{firewood}. In addition to the dialogue, the dataset also contains metadata about the arguments used by the negotiators. DealOrNoDeal involves three arbitrarily-defined issues: \textit{books}, \textit{hats}, and \textit{balls}. Our main goal is to perform opponent modeling for CaSiNo. To this end, we adapt DealOrNoDeal along with the available metadata in CaSiNo for data augmentation.

We refer to the CaSiNo Dialogues as CD, CaSiNo Argument metadata as CA, and DealOrNoDeal dialogue data as DND. While the CD data can be used as it is with our model, we adapt the other two data sources (CA and DND) to make them suitable for our approach (see Figure \ref{fig:architecture}). We now describe these adaptations.

\subsubsection{Capturing Preferences}
\label{sec:capture-preferences}
In order to provide more direct supervision for the preferences, we leverage the metadata from CaSiNo (CA data), where the participants explicitly reported their arguments for needing or not needing a specific issue (blue segment from Figure \ref{fig:architecture}). For instance, if food is the highest priority issue for a participant, they were asked to come up with an argument from their personal experiences as to why they would need food the most for camping.\footnote{These priority orders were randomly assigned to the participants by the authors of the CaSiNo paper.} Example arguments are provided in Figure~\ref{fig:architecture}. The participants came up with a variety of such arguments covering \textit{Personal Care}, \textit{Recreational}, \textit{Group Needs} or \textit{Emergency} requirements.\footnote{We refer the readers to the CaSiNo dataset paper for more examples around these themes.} The participants were then encouraged to leverage these arguments in their upcoming negotiations.

This metadata can provide more direct feedback on which implicit preference statements can lead to a higher or a lower affinity towards a specific issue. To incorporate this, we create dummy dialogues using templates and add them to the training data for our opponent modeling task. Consider a set of arguments $A$ = $(A_H, A_M, A_L)$, containing one argument for $H$igh, $M$edium, and $L$ow priorities respectively. We extract two pairs: $(A_H, A_L)$ and $(A_M, A_L)$ and construct the dummy dialogue as per Figure~\ref{fig:architecture}.\footnote{We skip the third pair due to an absence of a visible difference based on our qualitative analysis.} We ordered the arguments randomly to avoid any induced biases.

For each constructed dialogue, we only have ground-truth ranking order for a single pair of issues. Hence, the pairwise loss function from Equation~\ref{eq:margin-loss} needs a special treatment to ignore the score of the issue that is not relevant for a given dialogue. More specifically, while training with these constructed dialogues, we partially mask the margin ranking loss to only consider the loss from the pair for which the relation is known. Further, since a partial dialogue is not meaningful in this case, we only train the model with $\mathcal{L}_2$ loss using $k$=$2$.

Although we use the readily available metadata from CaSiNo in our work, we believe that such contextual data can be constructed for other realistic domains as well, such as by leveraging appropriate domain-specific knowledge about the negotiators' common requirements.

\subsubsection{Capturing Offers}
\label{sec:capture-offers}
To better capture the preferences in the previous section, our approach was to construct synthetic dialogues from a resource that primarily focused on implicit preference statements, so as to teach the model in a more explicit manner. With a similar idea, we adapt DND dialogues to better use the offer statements (green segment in Figure \ref{fig:architecture}). The DND dataset follows the same multi-issue framework as CaSiNo, which enables our adaptation. Each dialogue in DND involves three \textit{arbitrarily-defined} issues: \textit{books}, \textit{balls}, and \textit{hats}. Due to the arbitrary nature of these issues, there is minimal context discussed in the dialogues, reducing it to essentially an exchange of offers from both sides (see example in Figure~\ref{fig:architecture}). Hence, such a resource can be used to provide more explicit supervision to learn from the offer statements of the opponent. We map these dialogues to our dataset by \textit{randomly mapping the issues in this dataset to the issues in the target dataset}, in our case, CaSiNo. We modify the utterances by replacing all the occurrences of the issues with the corresponding issues in CaSiNo. For this purpose, we find that simple regular expressions prove to be effective (Appendix \ref{sec:dnd-regex-appendix}). Once mapped, this adapted data is simply added to the training data for our opponent modeling task.

\noindent\textbf{Note on multi-issue negotiations}: Our adaptation described above leverages the structural similarities between the two datasets. If the tasks follow a similar structure, it is relatively straightforward to use adaptations as described above for other settings as well. This can be largely done with regular expressions but if not, this relatedness still paves the way for multi-task learning. The negotiations in DealOrNoDeal and CaSiNo are based on a popular abstraction in the negotiation literature, referred to as the Multi-Issue Bargaining Task, or MIBT~\cite{fershtman1990importance}. MIBT is a generic framework that can be useful for many negotiation tasks beyond these datasets as well, for instance, salary negotiations, or negotiations between art collectors distributing the items among each other. It is extensively used in NLP~\cite{lewis2017deal,chawla2021casino,yamaguchi2021dialogue}, beyond NLP~\cite{mell2017grumpy}, and in the industry as well (e.g. iDecisionGames\footnote{\url{https://idecisiongames.com/promo-home}}).

\section{Experimental Design}
\label{sec:expt-design}
We address the following questions: \textbf{Q1) How useful is the proposed transformer-based ranker along with data augmentations for opponent modeling in negotiation dialogues?} We experiment with two pretrained language models and compare our ranker to standard baselines. To test the data augmentations, we analyze model ablations, including $0$-shot and few-shot settings. We also observe if they lead to a better performance with a lower number of utterances. \textbf{Q2) Do preferences and offers contribute to the performance?} To further shed light on the contributions of these utterances to the final opponent modeling performance, we look at average attention scores on these utterances. Further, for a more explicit analysis, we observe whether the performance varies by the \textit{integrative potential} in the negotiation, which essentially captures how aligned the preferences of the two negotiators are~\cite{chawla2021casino}. The scenarios with low integrative potential are usually associated with a higher expression of preferences and offers. Hence, we expected the performance to be higher in the cases with low integrative potential. \textbf{Q3) How does our approach compare to a human expert?} We compare our model to a human expert and recognize some of the errors that the model makes, discussing potential directions for future work.

\noindent\textbf{Datasets:} Each data point in CD results in \textit{two} dialogues for our analysis, based on the \textit{perspectives} of the two negotiators (Section \ref{sec:problem-formulation}). We report results on $5$-fold cross validation for this dataset. We further leave out $100$ dialogues from the training data for hyperparameter tuning, resulting in $1548$ dialogues for training, $100$ for tuning, and $412$ for evaluation - for each cross fold. We extract CA from the metadata corresponding to the training data of CD, leaving out $200$ constructed dialogues for validation (following Section \ref{sec:capture-preferences}). For DND data, we only select the dialogues with at least $4$ total utterances and unique priority values for meaningful training. After adaption (following Section \ref{sec:capture-offers}), we end up with $4074$ dialogues for training and $444$ for validation. All the models are primarily validated and tested on the corresponding subsets of CD (except for some additional analysis presented in Section \ref{sec:results}).

\noindent\textbf{Evaluation Metrics:} Our metrics are inspired by the negotiation literature, along with related research in Dialog State Tracking (DST) and Learning-to-Rank(LTR) tasks in NLP. Our primary metric is Exact Match Accuracy (EMA): the percentage of cases where the predicted priority order is entirely correct. This is analogous to the popular Joint Goal Accuracy in DST which captures the cases where all the slots are correctly identified~\cite{balaraman2021recent}. For negotiation tasks, even knowing the topmost priority can be useful. Hence, we also report Top-$1$ Accuracy: the percentage of cases where the highest priority issue is correctly predicted. Finally, we report the Normalized Discounted Cumulative Gain (NDCG@3). NDCG has been widely used in LTR tasks with distinct relevance values~\cite{yates2021pretrained}, which is also true for the setting that we consider. In our case, we use the relevance values as $5$, $4$, and $3$ for the most, second, and least ranked issues respectively, following the incentive design structure of CaSiNo. We compute these metrics for all $k$ from $1$ to $5$, varying the number of opponent utterances seen by the model. We present the results at $k$=$5$ to analyze the performance after seeing almost all of the opponent utterances in CaSiNo. To capture the performance with partial dialogues, we report corresponding $k$-penalty versions that take a weighted average of the performance for different values of $k$, while giving a linearly higher weight to the performance at a lower $k$.

\noindent\textbf{Methods:} We call the complete model from Figure \ref{fig:architecture} that combines all the three datasets for training as \textbf{CD + CA + DND}. We compare it with its ablations, including $0$-shot and few-shot scenarios. We further develop two standard baselines. The \textbf{Random} baseline chooses the final ranking at random, from all the possible orderings. \textbf{BoW-Ranker} is based on the Bag-of-Words paradigm. The input features are based on the normalized frequencies of the $500$ most frequent words in the training dataset, except stopwords. Instead of contextualized hierarchical representations, this method directly uses a feed-forward network on the input BoW features to predict the ranking. The model is trained on partial dialogues using the same margin ranking loss.

\noindent\textbf{Training Details:} The embedding dimension throughout is $768$ for transformer-based models. These models use base variant of either BERT~\cite{devlin2019bert} or RoBERTa~\cite{liu2019roberta} for Level I encoder. The Level II encoder uses one transformer layer. The feed-forward network contains two fully connected layers with a final sigmoid activation. We train the model with Adam optimizer using a learning rate of $2e^{-5}$ for transformer-based methods and $2e^{-3}$ for \textbf{BoW-Ranker}. The margin $c$ is kept as $0.3$. We use a dropout of $0.1$ to prevent overfitting. We further employ a loss-specific dropout of $0.15$, in order to backpropagate the loss from fewer $k$s simultaneously. The models were trained for $20$ epochs with a batch size of $25$. We checkpoint after every epoch and the one with the highest EMA at $k$=$5$ on the held out \textbf{CD} dataset is chosen for evaluation. We provide the details on the computing infrastructure, hyper-parameter tuning, and validation performance in Appendix \ref{sec:experiments-appendix}.

\begin{table*}[th]
\centering
\scalebox{0.7}{
\begin{tabular}{c|ccc|ccc}
\hline
\textbf{Model} & \multicolumn{3}{c|}{\textbf{$\mathbf{k}$=$\mathbf{5}$}} & \multicolumn{3}{c}{\textbf{$\mathbf{k}$-penalty}} \\
& \textbf{EMA} & \textbf{Top-$\mathbf{1}$}  & \textbf{NDCG$\mathbf{@3}$} & \textbf{EMA} & \textbf{Top-$\mathbf{1}$}  & \textbf{NDCG$\mathbf{@3}$} \\ \hline
\textbf{Random} & $16.46$ ($1.47$) & $32.49$ ($1.58$) & $48.49$ ($1.16$) & $16.59$ ($1.22$) & $33.99$ ($1.13$) & $49.76$ ($0.75$) \\
\textbf{BoW-Ranker} & $28.49$ ($1.3$) & $53.38$ ($2.21$) & $65.51$ ($0.62$) & $27.71$ ($1.24$) & $52.98$ ($1.97$) & $64.31$ ($1.67$) \\ \hline
\multicolumn{7}{c}{\textbf{Bert-based}} \\
\textbf{DND} & $41.12$ ($3.06$) & $64.69$ ($2.94$) & $73.88$ ($1.57$) & $34.5$ ($1.12$) & $58.75$ ($1.35$) & $68.48$ ($0.77$) \\
\textbf{CA+DND} & $41.9$ ($2.93$) & $66.98$ ($3.17$) & $75.91$ ($2.28$) & $36.01$ ($1.25$) & $61.09$ ($1.9$) & $70.09$ ($1.49$) \\
\textbf{CD} & $53.97$ ($3.02$) & $77.7$ ($2.85$) & $83.75$ ($1.96$) & $42.3$ ($1.53$) & $66.8$ ($1.78$) & $74.39$ ($1.45$) \\
\textbf{CD+CA} & $57.24$ ($3.09$) & $79.74$ ($2.37$) & $84.99$ ($1.87$) & $44.39$ ($1.17$) & $67.88$ ($1.16$) & $75.31$ ($1.1$) \\
\textbf{CD+DND} & $56.12$ ($4.07$) & $79.16$ ($2.57$) & $84.66$ ($1.84$) & $43.79$ ($2.07$) & $68.18$ ($1.55$) & $75.38$ ($1.6$) \\
\textbf{CD+CA+DND} & $56.56$ ($2.07$) & $80.13$ ($1.07$) & $85.49$ ($1.09$) & $44.22$ ($1.82$) & $69.21$ ($2.05$) & $76.03$ ($1.6$) \\ \hline
\multicolumn{7}{c}{\textbf{RoBerta-based}} \\
\textbf{DND} & $45.21$ ($3.07$) & $68.1$ ($2.8$) & $77.01$ ($1.76$) & $37.66$ ($1.41$) & $61.41$ ($2.3$) & $70.44$ ($1.5$) \\
\textbf{CA+DND} & $46.76$ ($1.89$) & $68.73$ ($1.22$) & $77.65$ ($0.9$) & $39.43$ ($1.67$) & $62.87$ ($2.5$) & $71.7$ ($1.83$) \\
\textbf{CD} & $60.06$ ($3.01$) & $81.98$ ($1.75$) & $86.54$ ($1.31$) & $46.57$ ($1.6$) & $69.26$ ($1.69$) & $76.17$ ($1.22$) \\
\textbf{CD+CA} & $60.01$ ($2.23$) & $80.23$ ($2.11$) & $85.85$ ($1.41$) & $46.96$ ($2.1$) & $68.59$ ($1.93$) & $76.05$ ($1.14$) \\
\textbf{CD+DND} & $62.54$ ($3.3$) & $82.56$ ($1.24$) & $\mathbf{87.57}$ ($1.18$) & $47.69$ ($2.52$) & $69.98$ ($1.96$) & $76.71$ ($1.55$) \\
\textbf{CD+CA+DND} & $\mathbf{63.57}$ ($3.44$) & $\mathbf{82.76}$ ($2.47$) & $87.55$ ($1.58$) & $\mathbf{48.72}$ ($2.03$) & $\mathbf{70.03}$ ($1.63$) & $\mathbf{77.14}$ ($1.38$) \\ \hline

\end{tabular}}
\caption{\label{tab:main-results}Performance on the opponent modeling task, showing the utility of the proposed methods. EMA and Top-$1$ represent the accuracy in percentage. We also scaled NDCG$@3$ to $0$-$100$. For all the metrics, higher is better. The numbers represent Mean (Std.) over $5$-cross folds of the \textbf{CD} data.}
\end{table*}

\section{Results and Discussion}
\label{sec:results}

\begin{figure*}[ht]
\centering
\begin{subfigure}[b]{0.4\textwidth}
 \centering
 \includegraphics[width=\textwidth]{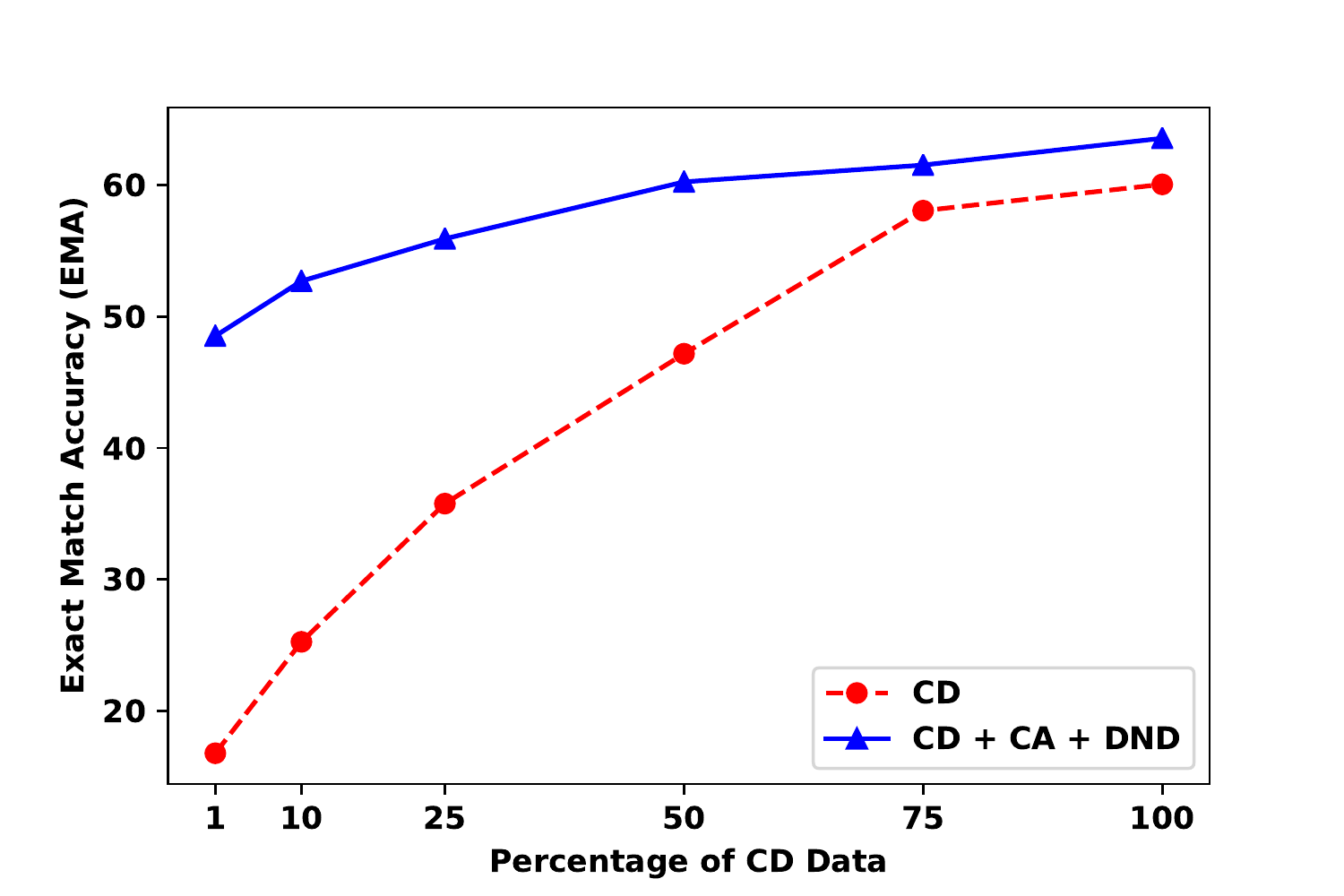}
 \caption{\label{fig:data-subsets}}
\end{subfigure}
\begin{subfigure}[b]{0.4\textwidth}
 \centering
 \includegraphics[width=\textwidth]{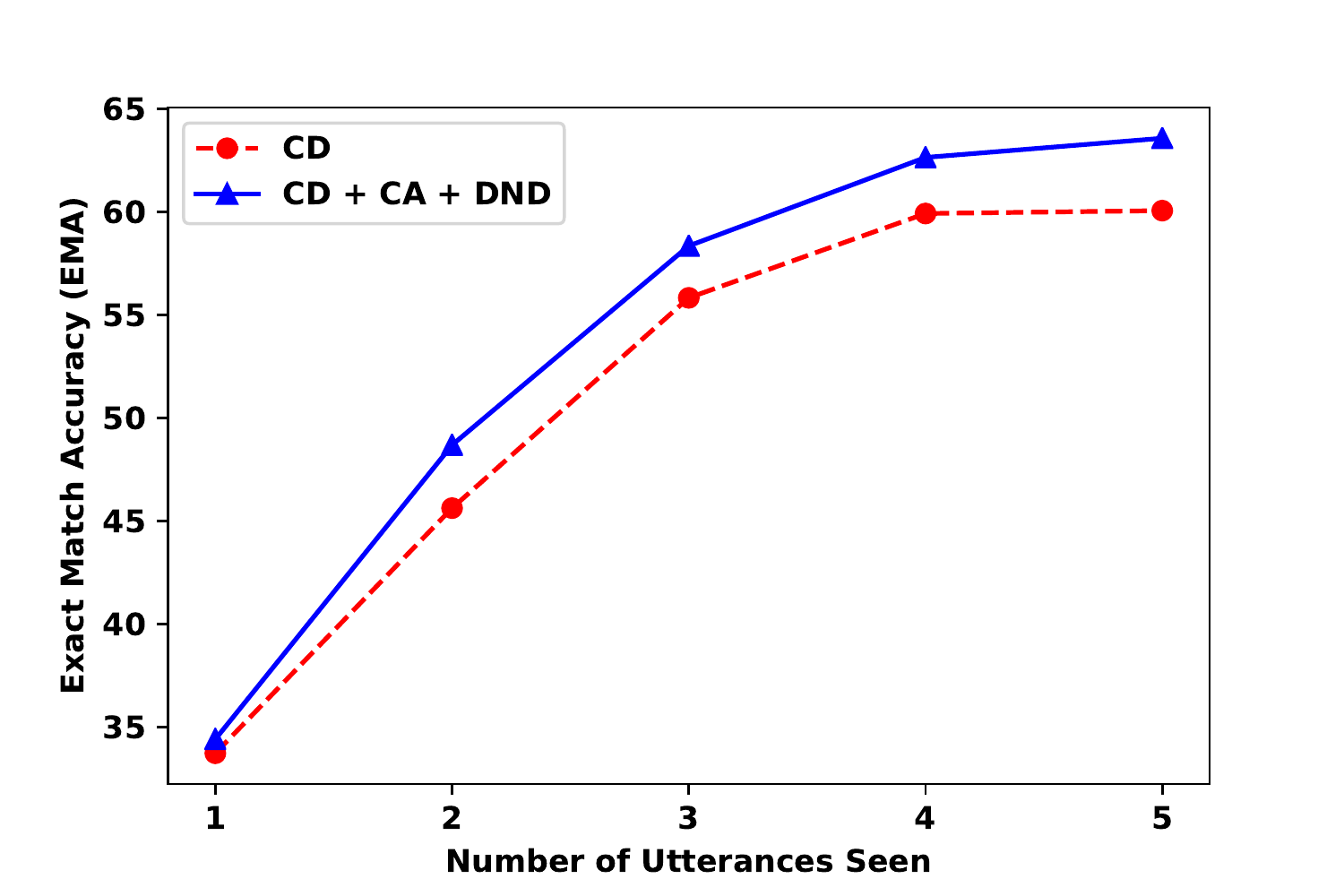}
 \caption{\label{fig:utterance-num}}
\end{subfigure}
\caption{Mean performance for two RoBERTa-based models: (a) on different percentages of \textbf{CD} data. The Y-Axis represents EMA at k=$5$, (b) on different values of $k$.}
\end{figure*}

\begin{table}[ht]
\begin{subtable}[h]{0.15\textwidth}
\centering
\scalebox{0.7}{
\begin{tabular}{c|c}
\multicolumn{2}{c}{\textbf{CA}} \\ \hline
\textbf{Model} & \textbf{Accuracy} \\ \hline
\textbf{Random} & $52.4$ ($4.14$) \\ \hline
\textbf{AD} & $63.8$ ($9.33$) \\
\textbf{AD+DND} & $73.4$ ($6.19$) \\
\textbf{CD+AD} & $78.9$ ($1.39$) \\
\textbf{CD+AD+DND} & $76.7$ ($3.52$) \\ \hline
\end{tabular}}
\caption{\label{tab:ca-results}}
\end{subtable}
\hspace{1cm}
\begin{subtable}[h]{0.15\textwidth}
\scalebox{0.7}{
\begin{tabular}{c|c}
\multicolumn{2}{c}{\textbf{DND}} \\ \hline
\textbf{Model} & \textbf{EMA} \\ \hline
\textbf{Random} & $16.04$ ($0.92$)\\ \hline
\textbf{DND} & $60.68$ ($2.05$)\\
\textbf{AD+DND} & $60.9$ ($1.87$) \\
\textbf{CD+DND} & $63.11$ ($1.77$) \\
\textbf{CD+AD+DND} & $63.56$ ($0.94$) \\ \hline
\end{tabular}}
\caption{\label{tab:dnd-results}}
\end{subtable}
\caption{Performance for RoBERTa-based models: (a) argument classification accuracy on the validation set of \textbf{CA}, (b) EMA at $k$=$2$ for opponent modeling on the validation set of \textbf{DND}. The numbers represent Mean (Std.) over $5$-cross folds.}
\end{table}

\subsection{Addressing Q1}

We summarize the results in Table \ref{tab:main-results}. Our proposed ranking-based models beat the \textbf{Random} and \textbf{BoW-Ranker} baselines by a huge margin across all metrics. This is true even for zero-shot \textbf{DND} and for \textbf{CA + DND}, attesting to the utility of the proposed ranking methodology and data adaptations.\footnote{Training with the \textbf{CA} data only was not useful due to the lack of training with any partial dialogues.} Comparing similar configurations, we observe that RoBERTa-based models outperform BERT-based models on this task. The best performing configuration is the RoBERTa \textbf{CD + CA + DND} that combines all the three data sources.

In Figure \ref{fig:data-subsets}, we plot the performance for different percentages of \textbf{CD} data. We only show RoBERTa-based models due to their superior performance. The plot highlights the advantage of adapting the related data sources, especially in few-shot settings, with \textbf{CD + CA + DND} at $50\%$ matching the performance of \textbf{CD} at $100\%$.

We also look at how the performance varies with the number of utterances seen in Figure \ref{fig:utterance-num}. We find that the performance gains are visible across all values of $k$. The data augmentations allow the model to perform better than the baselines, while observing a fewer number of utterances, making the model more useful in realistic scenarios.

\textbf{Performance on the adapted datasets:} We analyze if our joint learning also improves the performance on the validation sets of \textbf{CA} and \textbf{DND} datasets, showing advantages across multiple tasks. For \textbf{CA} dataset, we measure argument ranking accuracy: for a given input dialogue based on a pair of arguments, we consider a prediction as correct if the scores predicted by the model correctly rank the arguments. For \textbf{DND}, we analyze \textbf{EMA} at k=$2$ for opponent modeling, similar to our setup for CaSiNo. As evident from Tables \ref{tab:ca-results} and \ref{tab:dnd-results}, we find support that joint learning improves the performance on \textbf{CA} and \textbf{DND} datasets as well.

\subsection{Addressing Q2}

\textbf{Average attention:} We recognize the utterances with preference statements by utilizing strategy annotations in CaSiNo~\cite{chawla2021casino}. We assume that an utterance contains a preference if it was annotated with at least one of \textbf{Self-Need}, \textbf{Other-Need}, or \textbf{No-Need} strategies. For identifying offers, we use regular expressions following prior work~\cite{he2018decoupling} (refer Appendix \ref{sec:offer-regex-appendix}). We consider any utterance that is not labeled with a preference or an offer as \textit{Other}. Then, we observed the average attention put by the best-performing model on these categories in the Level II encoder. Preferences received an average of $0.3$, offers received $0.27$, and other utterances received $0.08$ attention scores, without any explicit indication about these categories during model training. We consider this as preliminary evidence that the learning process matches our intuition, with preferences and offers contributing to the performance.

\textbf{Performance across integrative potential:} For more concrete evidence of the utility of preferences and offers, we look at how the performance varies between scenarios with low and high integrative potential. This basically captures how aligned the preferences of the two negotiators are in a negotiation. In a scenario with low integrative potential, the negotiations are more competitive, leading to a higher expression of preferences and offers and providing a better signal to our ranking models. For our best-performing model, we find EMA at $k$=$5$ to be $68.75$ ($4.58$) for scenarios with low integrative potential against $60.31$ ($2.67$) for those with high potential. This provides stronger evidence that the learning process sensibly takes into account the preference and offer statements in the data.

\subsection{Addressing Q3}

\textbf{Comparison to Human Expert}: Similar to the trained models, we asked a human expert (an author of this work) to guess the priority order of the opponent by accessing partial dialogues. The expert was allowed to make multiple guesses if she is unsure, in which case the final ranking was chosen randomly from all the guesses. We compare the expert to our best-performing model on $100$ dialogues from the evaluation set. The expert achieved $75\%$ mean EMA at $k$=$5$ against $66\%$ for the model while performing better on other metrics as well. We show the comparison by varying the parameter $k$ in Appendix \ref{sec:human-appendix}.

While the model performs reasonably, there is a scope for improvement. We performed a qualitative analysis of the errors made by the model and the expert. In many cases, it is simply not feasible to predict accurately, especially when negotiators engage in small talk early on - indicating a limited scope for improvement with fewer utterances. In some cases, there is more focus on the highest priority issue, giving less explicit signals of the entire ranking. This might work for some applications but in other cases, the agent design can be modified to discuss the complete ranking more explicitly. Integrating other datasets that follow the same MIBT structure (such as~\cite{devault2015toward}) via data adaptation or multi-task learning is another potential direction. We also observed errors in the cases that included longer contextually-dense utterances, where preferences are shared indirectly as a response to the partner, and when the negotiators give away their higher priority issues out of empathy towards their partner. These cases are easier for the expert but can be confusing to the model. Better modeling of the prior context and handling of longer utterances are also avenues for improvements in the future.

\section{Related Work}
Opponent modeling encompasses several tasks in negotiations such as priority estimation, predicting opponent limits like BATNA~\cite{sebenius2017batna}, and classifying opponents into categories such as based on personality traits~\cite{albrecht2018autonomous, baarslag2016learning}. In this work, we focused only on inferring the opponent's priorities but in a more challenging domain involving chat-based interactions, instead of structured communication channels often used in prior work~\cite{williams2012iamhaggler, mell2017grumpy, johnson2021comparing}. Using a realistic interface like natural language fundamentally alters the negotiation dynamics in terms of the exchange of information, and hence, requires a separate investigation.

For chat-based negotiations,~\citet{nazari2015opponent} relied on heuristics and utterance-level annotations to infer the opponent's priorities using frequency-based methods. \citet{langlet2018detecting} explored a symbolic rule-based system to parse the utterances collected from a multimodal interaction. Instead, our focus is on modeling the priorities directly from partial dialogues as input. Research in negotiation dialogue systems has mainly focused on end-to-end modeling of the agent, without any explicit opponent modeling~\cite{lewis2017deal, he2018decoupling, zhou2019augmenting, cheng2019evaluating, parvaneh2019show}. However, there is evidence that even end-to-end systems can benefit from being more opponent-aware, as seen in recent work that uses dialogue acts to estimate opponent's behavior~\cite{zhang2020learning, yang2021improving}.

A number of related data augmentation strategies have been explored in Computer Vision and NLP~\cite{shorten2019survey, feng2021survey}. Most methods use rules or models to transform the available data or create synthetic data to avoid overfitting while training. This especially helps in low-resource languages~\cite{li2020diverse} and few-shot scenarios~\cite{kumar2019closer}.

\section{Conclusion}
We presented and evaluated a transformer-based approach for opponent modeling in negotiation dialogues. Our objective was to address the challenges to bridge the gap between existing research and practical applications of opponent modeling techniques. Our comparison to baselines and ablations attest to the utility of our method. We found that the proposed data adaptations can be especially beneficial in $0$-shot and few-shot scenarios. In the future, we will explore two primary directions: first, improving the model performance on opponent modeling by leveraging other related available datasets and by better incorporating the negotiation dialogue context, and secondly, using effective opponent modeling techniques towards the design of automated negotiation systems for applications in pedagogy and conversational AI.

\section{Broader Impact and Ethical Considerations}
\label{sec:broader-ethics}
\noindent\textbf{Datasets Used:} Both the datasets used in this work had been completely anonymized before their release by the respective authors. Moreover, we carefully verified the licensing details and ensured that the datasets were only used within the scope of their intended usage. 

We note that both datasets follow the multi-issue structure where the priority order remains fixed throughout the negotiation. Although this may not be true for some real-world scenarios, as we noted earlier, the underlying MIBT framework used by these datasets has been extensively used in academic research and also in the industry, attesting to the generalizability and applicability of this approach. Finally, we note that both the datasets are in English. Although this means that our experiments were limited to one language, our approach makes no such assumptions and should be broadly applicable to other settings as well. We encourage researchers to extend this work and study human-machine negotiations for other languages as well. This would open up exciting avenues for cross-culture research in this space, given the well-documented differences in how humans negotiate across cultures~\cite{luo2008analysis,andersen2018cultural}.

\noindent\textbf{Human Annotations:} Human annotations were used to estimate the expert performance on this task. This did not involve any additional crowdsourcing effort. Instead, the dialogues were annotated by an author of this work.

\noindent\textbf{Opponent Modeling For Negotiation Dialogues:} Negotiations are typically non-collaborative in nature, where the goals of the negotiating parties may not align with each other. Hence, the negotiators may not always feel comfortable in revealing their preferences for fear of being exploited. Even if they do, inferring them from natural language is challenging as preferences might be implied, and resolving these implications involves domain-specific knowledge and prior dialogue context. Regardless, incorporating such realistic communication channels is critical for designing practical and robust AI systems for downstream applications. However, most of the prior efforts in negotiations use restrictive menu-driven systems based on button clicks. Our work is a step towards bridging this gap.

This work is aligned with our broader goals for building automated negotiation systems, trained either in an end-to-end or a modular manner. For conversational AI applications, opponent modeling systems that can predict the priorities of the opponent reliably based on a partial dialogue can inform the strategy of the agent in the latter parts of the conversation. From the perspective of pedagogical applications, even the systems that can predict the priorities of a negotiator at the end of the negotiation can be helpful. For instance, consider a negotiation between two students, A and B who are asked to guess the opponent's priorities at the end of their negotiation. If the pedagogical agent is able to accurately guess the priorities of student B, while student A fails to guess correctly, this can be used to give concrete feedback to students who fail to recognize these strategies.

\noindent\textbf{Ethical Recommendations:} Finally, we briefly discuss the ethical considerations around the design of automated negotiation systems. A considerable amount of research in negotiations has focused on ethics. Primary concerns revolve around the acts of emotion manipulation, bias, deception, and misinterpretation~\cite{lewicki2016essentials}. Consequently, these issues can also emerge in the systems that are developed on human-human negotiation dialogue datasets. Our central recommendation in mitigating the impact of these issues for negotiation dialogue systems or other conversational AI assistants is transparency - around the identity, capabilities, and any known undesirable behaviors of the system. Further, any data collected during the deployment phase should be properly anonymized and the users of the system should be well-informed. In particular, we recommend extra precautions for systems that are adaptive towards their opponents or users such as having regular monitoring for any unexpected behaviors, to ensure that the systems are not offensive or discriminatory.

\section*{Acknowledgements}
We would like to thank Garima Rawat, Thamme Gowda, Sarik Ghazarian, Abhilasha Sancheti, and Prakhar Gupta for their valuable comments. We also thank the anonymous reviewers for their valuable time and feedback. Our research was sponsored by the Army Research Office and was accomplished under Cooperative Agreement Number W911NF-20-2-0053. The views and conclusions contained in this document are those of the authors and should not be interpreted as representing the official policies, either expressed or implied, of the Army Research Office or the U.S. Government. The U.S. Government is authorized to reproduce and distribute reprints for Government purposes notwithstanding any copyright notation herein.

\bibliography{anthology,custom}
\bibliographystyle{acl_natbib}

\clearpage

\appendix

\section{Experiments}
\label{sec:experiments-appendix}
\subsection{Computing Infrastructure}
All experiments were performed on a single Tesla V100 GPU. The complete model (\textbf{CD + CA + DND}) takes around $10$ hours for training with $32$-bit precision on a single cross-validation fold with a batch size of $25$.
\subsection{Training Details}
We used a combination of randomized and manual search to tune the hyperparameters. For each cross fold, we kept $50$ dialogues from the CD training data for parameter tuning. This amounts to $100$ data points, considering the two perspectives extracted from each dialogue. The metric for choosing the best hyperparameters is EMA at $k$=$5$, averaged over the $5$ cross-validation folds. We tuned the parameters on the performance of the BERT-based model with \textbf{CD + CA + DND} configuration.

We vary the learning rate in $\{1e^{-5}$, $2e^{-5}$, $3e^{-5}\}$, dropout in $\{0.0$, $0.1$, $0.2\}$, and loss-specific dropout in $\{0.0$, $0.15$, $0.25\}$. We also varied the number of transformer layers in Level II encoder from Figure \ref{fig:architecture} in the set $\{1$, $2$, $3\}$. For DND, we also varied the number of instances that were chosen for adaptation but found that using all the instances that passed our filtering gave the best performance. We further varied the margin for ranking loss in $\{0.0$, $0.3$, $0.5\}$. Finally, for the models trained on combined datasets, we tried with a higher weightage ($2$x) for the loss contribution of CA-adapted instances due to their lower total count but found no visible improvements in the performance. The rest of the hyper-parameters were fixed based on the available computational and space resources. We report the best performing hyper-parameters in the main paper.

The models used in the paper have nearly $171$ million trainable parameters. We report the mean performance on the validation set in Table \ref{tab:val-results}.

\begin{table}[th]
\centering
\scalebox{0.9}{
\begin{tabular}{c|c}
\hline
\textbf{Model} & \textbf{EMA} \\ \hline
\textbf{Random} & $17.8$ ($4.87$) \\
\textbf{BoW-Ranker} & $35$ ($3.35$) \\ \hline
\multicolumn{2}{c}{\textbf{Bert-based}} \\
\textbf{DND} & $51$ ($1.67$) \\
\textbf{CA + DND} & $51.2$ ($3.12$)\\
\textbf{CD} & $63.6$ ($4.84$)\\
\textbf{CD + CA} & $65.8$ ($1.94$) \\
\textbf{CD + DND} & $69$ ($2.28$)\\
\textbf{CD + CA + DND} & $70$ ($2.61$)\\ \hline
\multicolumn{2}{c}{\textbf{RoBerta-based}} \\
\textbf{DND} & $54.6$ ($5.43$)\\
\textbf{CA + DND} & $55$ ($5.55$)\\
\textbf{CD} & $70.2$ ($3.19$)\\
\textbf{CD + CA} & $70$ ($3.95$)\\
\textbf{CD + DND} & $75.6$ ($2.15$)\\
\textbf{CD + CA + DND} & $\mathbf{77.8}$ ($2.32$)\\ \hline
\end{tabular}}
\caption{\label{tab:val-results}Validation performance for opponent modeling on CD dataset. The reported EMA is at $k$=$5$. The numbers represent Mean (Std.) over $5$-cross folds of the CD data.}
\end{table}

\subsection{External Packages and Frameworks}
The models were developed in PyTorch Lightning\footnote{\url{https://www.pytorchlightning.ai/}} and relied on the HuggingFace Transformers library\footnote{\url{https://github.com/huggingface/transformers}} for using the pretrained models and their corresponding tokenizers. We used a number of external packages such as Python Scikit Learn\footnote{\url{https://scikit-learn.org/stable/modules/model_evaluation.html}} library for implementing the evaluation metrics, and NLTK\footnote{\url{https://www.nltk.org/api/nltk.tokenize.html}} for tokenization for the Bag-of-Words model.

\section{Regular Expression Usage}

\subsection{Adapting DealOrNoDeal data}
\label{sec:dnd-regex-appendix}
We randomly mapped \textit{book} from DealOrNoDeal to \textit{food}, replacing all occurrences of `book' and `books' with `food' in the utterances. Similarly, \textit{hat} was mapped to \textit{water}, and \textit{ball} was mapped to \textit{firewood}. Since the dialogues only involve minimal context about the issues, we found these replacements to be sufficient.

\subsection{Identifying Offer statements}
\label{sec:offer-regex-appendix}

The offer statements were also recognized by regular expressions for the purpose of computing average attention scores. Specifically, an utterance is classified as having an offer, if it contains $3$ or more of the following phrases - \{'0', '1', '2', '3', 'one', 'two', 'three', 'all the', 'food', 'water', 'firewood', 'i get', 'you get', 'what if', 'i take', 'you can take', 'can do'\}. The threshold $3$ and these phrases were chosen heuristically via qualitative analysis.

\section{Comparison with Human Performance}
\label{sec:human-appendix}

\begin{figure}[th]
\centering
 \includegraphics[width=\linewidth]{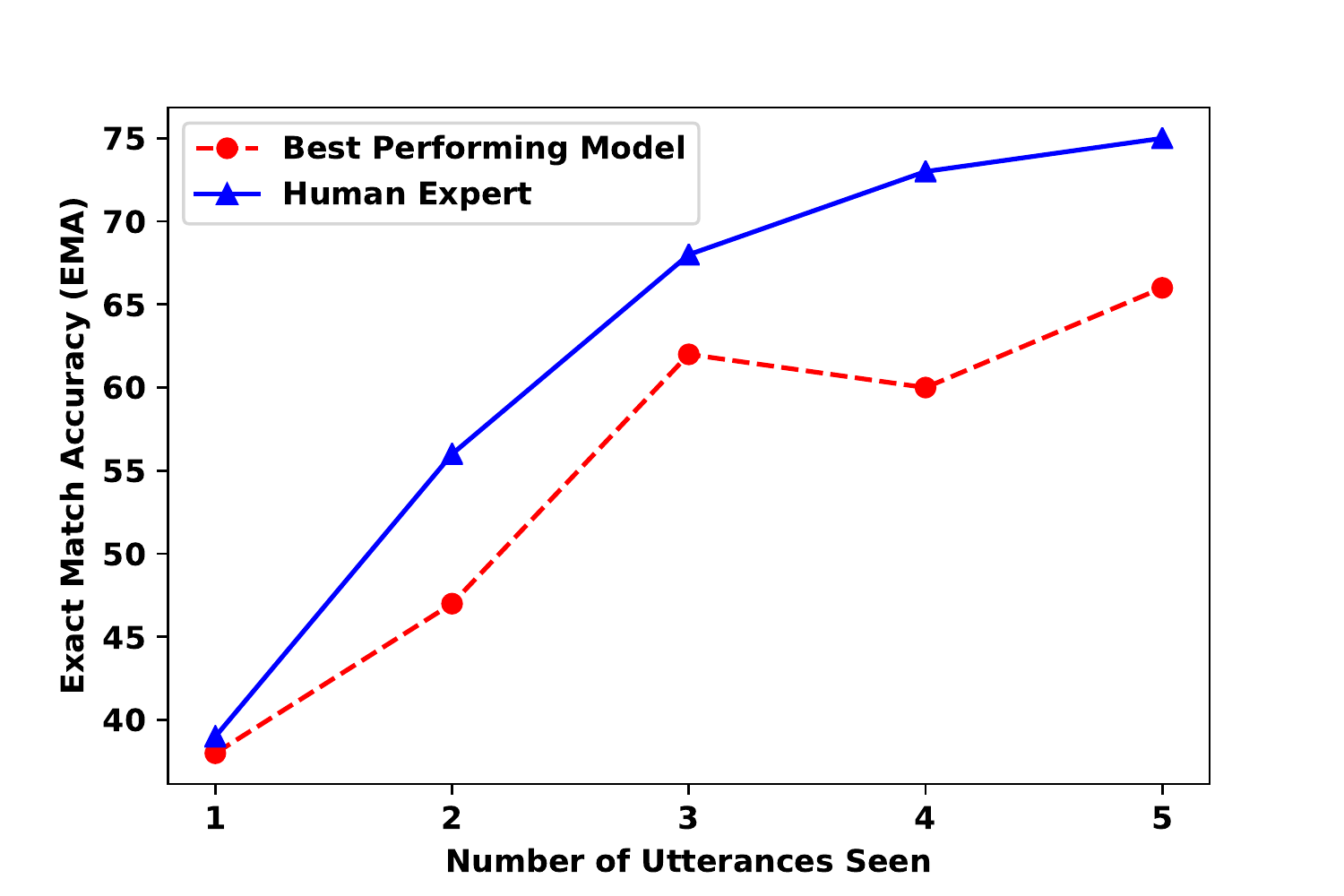}
\caption{Mean performance comparison for the best performing model with the human expert for different values of $k$.}
\label{fig:human-expert}
\end{figure}

We present the performance for our best performing model with the human expert across different values of $k$ in Figure \ref{fig:human-expert}.

\end{document}